\documentclass{article}
\usepackage{spconf,amsmath,graphicx,booktabs,multirow,bm,hyperref}
\usepackage[numbers]{natbib}
\pdfoutput=1

\makeatletter
\def\@sect#1#2#3#4#5#6[#7]#8{%
  \ifnum #2>\c@secnumdepth
    \let\@svsec\@empty
  \else
    \refstepcounter{#1}%
    \protected@edef\@svsec{\@seccntformat{#1}\relax}%
  \fi
  \@tempskipa #5\relax
  \ifdim \@tempskipa>\z@
    \begingroup
      #6{%
        \@hangfrom{\hskip #3\relax\@svsec}%
          \interlinepenalty \@M
          \ifnum #2=1\relax\bf\centering\uppercase{#8}\par%
          \else\ifnum #2=2\relax\bf #8\par%
          \else\it #8\par\fi\fi}%
    \endgroup
    \csname #1mark\endcsname{#7}%
    \addcontentsline{toc}{#1}{%
      \ifnum #2>\c@secnumdepth \else
        \protect\numberline{\csname the#1\endcsname}%
      \fi
      #7}%
  \else
    \def\@svsechd{%
      #6{\hskip #3\relax
      \@svsec \ifnum #2=1\relax\bf\uppercase{#8}\else#8\fi}%
      \csname #1mark\endcsname{#7}%
      \addcontentsline{toc}{#1}{%
        \ifnum #2>\c@secnumdepth \else
          \protect\numberline{\csname the#1\endcsname}%
        \fi
        #7}}%
  \fi
  \@xsect{#5}}
\makeatother

\title{PatMT2: Two-Stage Domain Adaptation for Multilingual Patent Translation}

\name{Xiaowei Su$^{1,2}$ \qquad Xusen Cheng$^{1,*}$}
\address{$^{1}$Renmin University of China, Beijing, China \\
  $^{2}$Beijing Bayuegua Technology Co., Ltd., Beijing, China \\
  \{xwsu, xusen.cheng\}@ruc.edu.cn \\
  $^{*}$Corresponding author \\
  ORCID: 0000-0002-7784-7141 (Su), 0000-0002-2028-1337 (Cheng)}

\begin{document}
\ninept
\maketitle

\begin{abstract}
We present PatMT2 (Patent Machine Translation, v2), a two-stage LoRA pipeline that adapts the Hy-MT2-7B dedicated translation model to four-language (Chinese/English/Japanese/Korean) patent translation. On six translation directions, the Stage-2 checkpoint reaches a macro-average BLEU of 43.39 (BF16) to 43.57 (AWQ INT4), improving over the Hy-MT2-7B base by up to +8.8 BLEU and also exceeding Qwen3-32B and several external systems on the same average. We then measure the cost of adaptation: probing 38 languages absent from the fine-tuning data, the model still produces correct, fluent output on 34, while four low-resource languages collapse into distinct, reproducible failure modes. We treat these as diagnostic cases rather than retention rates, and report their mechanisms without attributing them to a single cause. Paired bootstrap tests confirm the translation gains over all baselines ($p < 0.001$).
\end{abstract}


\section{Introduction}
\label{sec:intro}

Domain adaptation via supervised fine-tuning (SFT) is standard practice for specialized translation tasks, but its effect on a model's broader multilingual capabilities is rarely measured directly~\cite{luo2023empirical}. When a translation model pretrained on dozens of languages is fine-tuned on a narrow four-language corpus, does it retain its original breadth? The answer depends on the base model's architecture, capacity, and pretraining strategy---factors typically confounded in single-model studies.

This question matters for patent translation, where models must handle dense, legally load-bearing technical prose across multiple jurisdictions. A dedicated translation model like Hy-MT2-7B is pretrained specifically for translation; its multilingual competence is the product itself. A generalist LLM like Qwen3-32B~\cite{qwen2025qwen3} acquires translation ability as a side effect of broad pretraining. It is not obvious \emph{a priori} which type is more robust to narrow-domain fine-tuning.

\begin{figure}[t]
\centering
\includegraphics[width=\columnwidth]{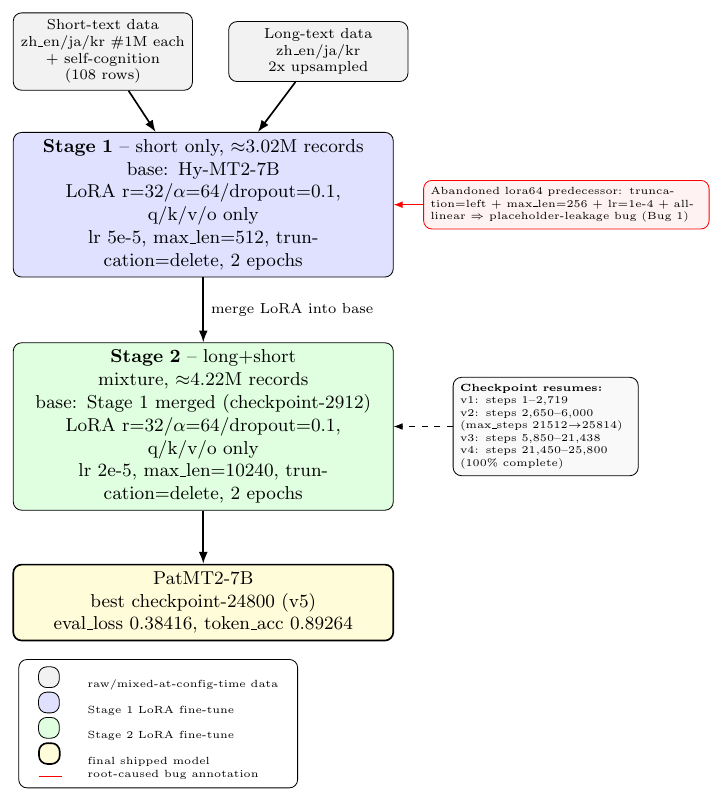}
\caption{Two-stage fine-tuning pipeline for PatMT2-7B. Stage~1 trains on short-text data; Stage~2 adds long-text mixed with skill-retention short-text. Both stages use attention-only LoRA adapters on the Hy-MT2-7B base.}
\label{fig:pipeline}
\end{figure}

We address this through PatMT2, a two-stage LoRA~\cite{hu2022lora} adaptation of Hy-MT2-7B for Chinese--English/Japanese/Korean patent translation (Fig.~\ref{fig:pipeline}). The pretrained PatMT2 model is currently being organized for public release and will be made available at https://huggingface.co/SuStarSmile/PatMT2. Our contributions are:
\begin{enumerate}
\item \textbf{The effectiveness and the boundary of patent-domain adaptation.} Starting from Hy-MT2-7B, we compare the base model against the Stage-1 (short-text) and Stage-2 (short+long) checkpoints across six directions and two length regimes, and report where the gained translation quality holds and where it does not. Two-stage LoRA is itself mature; the value here is a controlled adaptation question with reusable conclusions about when the recipe pays off.
\item \textbf{What the training stages do to multilingual capability.} On languages the base model supports but the fine-tuning data never covered, we measure quality, target-language correctness, repetition degradation, and stopping anomalies from the base model through each checkpoint, separating pre-existing base errors from errors introduced by adaptation.
\item \textbf{A validated training choice.} We compare the two-stage curriculum against a single-stage mixed run under a fixed data and token budget, and report the conditions and costs under which the two-stage ordering is preferred.
\end{enumerate}

\section{Related Work}
\label{sec:related}

\paragraph{Parameter-efficient fine-tuning.} LoRA~\cite{hu2022lora} freezes pretrained weights and injects trainable low-rank update matrices into linear projections. QLoRA~\cite{dettmers2023qlora} combines 4-bit quantization with LoRA adapters. We additionally show that the choice of \emph{which} linear layers receive adapters (attention-only vs.\ all-linear) has a measurable effect on multilingual retention.

\paragraph{Catastrophic forgetting.} Luo et al.~\cite{luo2023empirical} study catastrophic forgetting during continual instruction tuning of 1B--7B LLMs. Our study is complementary in scope but narrower in claim: we hold the fine-tuning task fixed and track one dedicated translation model through two adaptation stages, measuring what changes on languages outside the fine-tuning data. Because the checkpoints we compare differ in training data and in token budget as well as in ordering, we do not claim to isolate capacity or specialization as a causal factor; we report the observed changes and the conditions under which they occur.

\paragraph{Multilingual MT.} NLLB~\cite{nllb2024scaling} demonstrates large-scale multilingual machine translation across 200 languages. Our study examines how narrow four-language fine-tuning affects translation capabilities in languages covered by the base model but absent from the fine-tuning data.

\section{Two-Stage Adaptation Pipeline}
\label{sec:pipeline}

\subsection{Training Data}

The training corpus covers three language pairs (zh$\leftrightarrow$en, zh$\leftrightarrow$ja, zh$\leftrightarrow$kr). Stage~1 uses ${\approx}3$M short-text records (1M per pair); Stage~2 mixes ${\approx}1.8$M long-text records (upsampled 2$\times$) with 200K short-text skill-retention records, totaling ${\approx}4.2$M records. Records are mixed at training time via inline resampling, never physically merged.

\subsection{Training Configuration}

Table~\ref{tab:config} summarizes the adopted hyperparameters. Both stages use LoRA rank~32, alpha~64, dropout~0.1, targeting attention projections only (q/k/v/o\_proj), with DeepSpeed ZeRO Stage~2~\cite{rajbhandari2020zero}.

\begin{table}[t]
\centering
\caption{Two-stage training configuration.}
\label{tab:config}
\footnotesize
\setlength{\tabcolsep}{3pt}
\begin{tabular}{@{}lcc@{}}
\toprule
Parameter & Stage 1 & Stage 2 \\
\midrule
Base model & Hy-MT2-7B & Stg1 merged \\
LoRA rank/alpha & 32/64 & 32/64 \\
Dropout & 0.1 & 0.1 \\
Target modules & q/k/v/o & q/k/v/o \\
Learning rate & $5{\times}10^{-5}$ & $2{\times}10^{-5}$ \\
Schedule & cosine & cosine \\
Warmup ratio & 0.03 & 0.05 \\
Epochs & 2 & 2 \\
Max seq length & 512 & 10,240 \\
Truncation & delete & delete \\
Eff.\ batch size & 2,048 & 320 \\
Distributed & ZeRO-2 & ZeRO-2 \\
Records & ${\approx}3.0$M & ${\approx}4.2$M \\
Best ckpt & ckpt-2912 & ckpt-24800 \\
\bottomrule
\end{tabular}
\end{table}

An earlier lora64 recipe (rank~64, all-linear targeting, lr~$1{\times}10^{-4}$, left truncation, hardcoded system prompt) was abandoned after diagnosis revealed two compounding issues: (1)~left-truncation leaked chat-template placeholder tokens into training examples, corrupting 8.8\% of samples at max\_length=256; (2)~narrow data coverage, elevated learning rate, all-linear LoRA, and a language-limiting system prompt accelerated multilingual forgetting. The adopted lora32 configuration addresses causes (2)--(4); cause~(1) was resolved by switching to delete-truncation and retraining Stage~1 from scratch. Narrow data coverage was deliberately not addressed: no multilingual replay data was mixed.

\section{Evaluation}
\label{sec:eval}

\subsection{Translation Quality}

Table~\ref{tab:quality} reports BLEU scores for PatMT2-7B alongside baselines on a stratified benchmark (4,000 pairs/direction, short+long combined), using sacrebleu with language-appropriate tokenization. Quantized variants use FP8 and AWQ INT4~\cite{lin2024awq}.

\begin{table}[t]
\centering
\caption{BLEU across six directions (4,000 pairs/dir). Bold = column best.}
\label{tab:quality}
\footnotesize
\setlength{\tabcolsep}{2.5pt}
\begin{tabular}{@{}lccccccc@{}}
\toprule
Model & z$\to$e & e$\to$z & z$\to$j & j$\to$z & z$\to$k & k$\to$z & Avg \\
\midrule
HunYuan-MT-Chimera & 34.19 & 32.73 & 32.16 & 33.46 & 28.28 & 30.13 & 31.83 \\
Hy-MT2-7B (base) & 36.67 & 32.87 & 35.23 & 37.46 & 30.49 & 34.79 & 34.58 \\
Qwen3-32B (base) & 39.56 & 37.72 & 38.23 & 45.59 & 32.67 & 39.60 & 38.90 \\
PatMT-v1 (BF16) & 36.83 & 39.42 & 41.32 & 46.70 & \textbf{38.69} & 43.65 & 41.10 \\
Qwen3.6-27B-FP8 & \textbf{41.07} & 39.84 & 41.16 & \textbf{49.54} & 35.69 & 39.25 & 41.09 \\
\midrule
PatMT2 BF16 & 40.31 & 42.96 & 48.65 & 48.38 & 36.44 & 43.59 & 43.39 \\
PatMT2 FP8 & 40.18 & 42.70 & \textbf{48.69} & 43.95 & 36.11 & 43.43 & 42.51 \\
PatMT2 AWQ4 & 40.10 & \textbf{43.64} & 48.65 & 48.45 & 36.73 & \textbf{43.86} & \textbf{43.57} \\
\bottomrule
\end{tabular}
\end{table}

PatMT2-7B attains the highest six-direction average among all compared systems, ranging from 43.39 BLEU (BF16) to 43.57 BLEU (AWQ INT4, the single best average in Table~\ref{tab:quality}). Taking the BF16 checkpoint as the reference deployment, this is +8.8 BLEU over the Hy-MT2-7B base and +4.5 BLEU over Qwen3-32B. The average is a macro-average over per-direction BLEU, not a single pooled corpus BLEU; we report it alongside the per-direction columns so that a reader can see which directions actually drive it. No single PatMT2 variant wins every column: the per-column best is split across Qwen3.6-27B-FP8 (zh$\to$en, ja$\to$zh), PatMT-v1 (zh$\to$ko), and the PatMT2 FP8/AWQ4 variants (zh$\to$ja, en$\to$zh, ko$\to$zh); the BF16 checkpoint we highlight elsewhere in the paper is not the column-best on any individual direction.

The gains are not uniform, and we report the disagreements rather than claiming a clean sweep. Against the Hy-MT2-7B base, PatMT2-7B (BF16) wins on all six directions, though the margin varies from +3.6 on zh$\to$en to +13.4 on zh$\to$ja. Against the stronger external systems the picture is mixed rather than dominant. \texttt{Qwen3.6-27B-FP8} still leads on zh$\to$en (41.07 vs.\ 40.31) and on ja$\to$zh (49.54 vs.\ 48.38), while PatMT2-7B (BF16) leads on the remaining four directions. PatMT-v1, the earlier large domain-adapted model, is beaten by PatMT2-7B (BF16) on four directions but retains a small edge on zh$\to$ko (38.69 vs.\ 36.44) and on ko$\to$zh (43.65 vs.\ 43.59). We therefore claim the best six-direction average, not per-direction dominance.

Among quantized variants, AWQ INT4 marginally exceeds BF16 (43.57 vs.\ 43.39), while FP8 shows a small degradation (42.51). Direction-specific variation is notable: FP8 matches BF16 on zh$\to$ja but trails by 4.4 BLEU on ja$\to$zh, which is larger than the aggregate quantization gap and would be invisible in a single averaged number.

\subsection{Statistical Significance}

Paired bootstrap resampling (10,000 iterations, aligned by sample ID) confirms: PatMT2 vs.\ Hy-MT2 base $\Delta$BLEU = +8.8 ($p < 0.001$, Cohen's $d = 0.96$); vs.\ Qwen3-32B $\Delta$BLEU = +4.5 ($p < 0.001$, $d = 0.67$); vs.\ PatMT-v1 $\Delta$BLEU = +2.4 ($p < 0.001$, $d = 0.18$). Quantization effects are not significant: FP8 vs.\ BF16 $\Delta$BLEU = $-$0.088 ($p = 0.875$); AWQ vs.\ BF16 $\Delta$BLEU = +0.18 ($p > 0.05$).

\subsection{Multilingual Retention}
\label{sec:retention}

We probed PatMT2-7B (BF16) on 38 languages never included in the fine-tuning data, issuing one fixed source sentence per language and reading the raw output. This is a \emph{diagnostic probe}, not a language-level retention benchmark: a single sentence per language can surface a failure mode but cannot establish a retention rate or an accuracy figure. With that limit stated, direct reading of the raw outputs shows 34 of 38 languages producing correct, fluent output, including low-resource and non-Latin-script languages (Khmer, Burmese, Gujarati, Urdu, Hebrew, Kazakh, Persian). A purely automated non-empty-response check would report 38/38; the four failures below were found only by reading the text.

\textbf{Four languages fail} with distinct, reproducible mechanisms: Tibetan (repetition loop, 18K characters), Mongolian (single-syllable token loop), Tamil (one word repeated 1{,}949$\times$), and Uyghur (instruction echo, 45 characters, where the model translates its own instruction instead of the source). Three of the four are repetition collapse; Uyghur is qualitatively different and resembles an instruction-following failure rather than a translation-quality failure.

Cross-checking quantized variants shows non-monotonic behavior: AutoRound INT4 recovers 3 of the 4 collapse cases (37/38) but introduces a Gujarati failure, while AWQ INT4 recovers 2 of 4 (36/38) but fails differently on Tibetan and Uyghur. Quantization therefore interacts with fragile low-resource representations rather than uniformly helping or hurting. Because each language was probed with a single sentence, we treat all of these as fault cases to be characterized, not as measured retention rates, and we do not attribute them to one cause: the fine-tuning data, the base model's existing weaknesses, and the decoding configuration are all entangled in these observations.

\begin{figure}[t]
\centering
\includegraphics[width=\columnwidth]{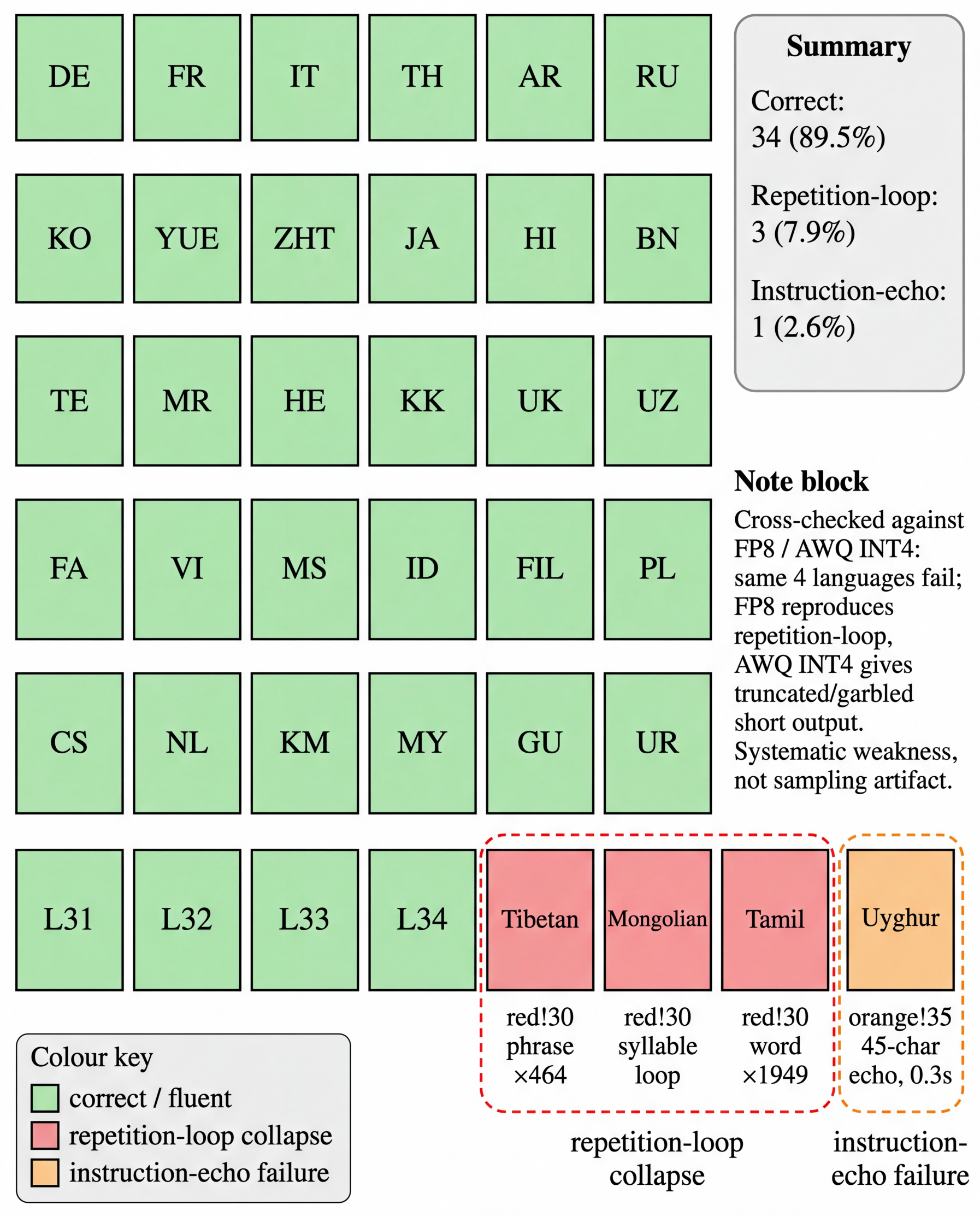}
\caption{Multilingual retention probe: 34 of 38 never-fine-tuned languages produce correct output, and 4 fail with distinct mechanisms. One fixed source sentence per language; this figure is a fault-case map, not a retention-rate estimate.}
\label{fig:retention}
\end{figure}

\section{Discussion and Conclusion}
\label{sec:conclusion}

Two-stage LoRA adaptation of Hy-MT2-7B yields a substantial and statistically validated patent translation gain: +8.8 BLEU on the six-direction average, with the largest margin on the zh$\to$ja direction and no direction where the adapted model falls below its own base. The four languages that fail the retention probe share low-resource profiles, and their failures are consistent with narrow training-data coverage, but our evidence does not isolate that as the cause.

We had originally intended to compare a 7B specialist against a 32B generalist to argue that capacity drives retention. Our own design does not support that claim, and we do not make it. The two systems differ in model family, pretraining, scale, training data, short/long text mix, and recipe, so any difference between them is not attributable to a single factor. Capacity and model specialization remain plausible hypotheses for follow-up, not findings of this work.

An earlier configuration (lora64: rank~64, all-linear targeting, higher learning rate, left truncation, language-limiting system prompt) was replaced by the lora32 recipe reported above. That change bundle altered rank, learning rate, target modules, truncation, and prompt simultaneously, so we present it as experience rather than as a single-factor ablation.

Limitations: (1)~the retention probe uses one sentence per language and cannot support retention rates; (2)~no multilingual replay data was mixed during training; (3)~only automatic metrics, no human evaluation; (4)~the v1 data split lacks family-level grouping, so contamination cannot be fully excluded. Future work: a multi-sample per-language retention study with reliable references, a single-stage mixed-training comparison under a fixed token budget, and human evaluation.

\bibliographystyle{IEEEbib}
\bibliography{references}

\end{document}